\documentclass[letterpaper, 10 pt, conference]{ieeeconf}

\usepackage{times}
\usepackage{epsfig}
\usepackage{graphicx}
\usepackage{amsmath}
\usepackage{amssymb}
\usepackage{float}
\usepackage{capt-of,etoolbox}

\usepackage[breaklinks=true,bookmarks=false]{hyperref}

\def\cvprPaperID{****} % *** Enter the CVPR Paper ID here


\begin{document}

% \newcommand\myfigure{%

% }

%%%%%%%%% TITLE
\title{6-DoF Grasp Planning using Fast 3D Reconstruction and Grasp Quality CNN}

\author{Yahav Avigal$^*$\\
UC Berkeley\\
yahav\_avigal@berkeley.edu\\
\and
Samuel Paradis$^*$\\
UC Berkeley\\
samparadis@berkeley.edu\\
\and
Harry Zhang$^*$\\
UC Berkeley\\
harryhzhang@berkeley.edu
}

\maketitle
\begin{figure*}
\includegraphics[width=0.9\textwidth]{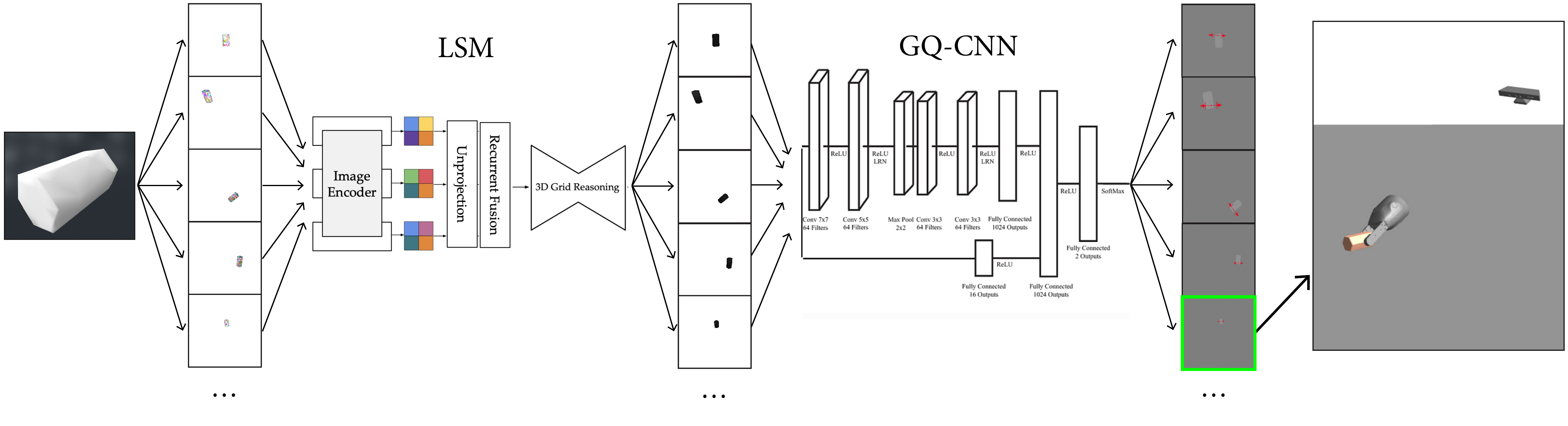} 
   \captionof{figure}{\textbf{Scene to Grasp Pipeline:} Starting with some graspable object, we first render multiple views of the scene from several cameras. Next, we use these renderings as input to an specially-trained LSM, which outputs depth maps. These depth maps are fed into a multi-view GQ-CNN, which generates the optimal grasp across all views.}

\label{fig:short}
\end{figure*}

%\thispagestyle{empty}
% \setcounter{figure}{1}

%%%%%%%%% ABSTRACT

\begin{abstract}
   Recent consumer demand for home robots has accelerated performance of robotic grasping. However, a key component of the perception pipeline, the depth camera, is still expensive and inaccessible to most consumers. In addition, grasp planning has significantly improved recently, by leveraging large datasets and cloud robotics, and by limiting the state and action space to top-down grasps with 4 degrees of freedom (DoF). By leveraging multi-view geometry of the object using inexpensive equipment such as off-the-shelf RGB cameras and state-of-the-art algorithm such as Learn Stereo Machine (LSM), the robot is able to generate more robust grasps from different angles with 6-DoF. In this paper, we present a modification of LSM to graspable objects, evaluate the grasps, and develop a 6-DoF grasp planner based on Grasp-Quality CNN (GQ-CNN) that exploits multiple camera views to plan a robust grasp, even in the absence of a possible top-down grasp.
\end{abstract}

%%%%%%%%% BODY TEXT

\section{Introduction}

As industrial robots become ubiquitous, the yearn for home robotic platforms that will automate mundane chores, such as folding laundry or getting the house in order, increases as well. With recent advances in speeding up imaging and grasp analysis, more advanced and accurate robotic platforms with expensive, high-resolution depth cameras have shown they can utilize data-driven approaches and generalize to grasp unseen objects when applying 4-DoF grasping, where the gripper is forced to approach objects in a top-down manner such that the planned grasp is parallel to the depth image plane. However, the home environment is more complex than the common structured industrial setting, and top-down grasp planning often fails to find a good grasp in clutter and unstructured environments where the information of the objects’ locations and geometries are partial and limited. Furthermore, the high cost of a top-tier depth camera makes it inaccessible to most households. 

This motivates the development of robust and economical systems that can leverage state-of-the-art algorithms using inexpensive hardware. One such development is eliminating the need for expensive top-down depth cameras, instead utilizing multiple RGB images from a variety of view points. Beyond decreasing cost, this can reveal more geometry and topology of the objects, further facilitating the possibility for 6-DoF grasping tasks. In order to conduct online grasping, we need a fast model that can generate depth maps from RGB input images in subsecond time, and that is invariant to the camera angle and location in the environment. One such architecture that fits this requirement is a Learnt Stereo Machine, or LSM. To generate grasps, we suggest the use of a GQ-CNN, which is a state-of-the-art architecture that generates a grasp from a top-down view depth map. However, to leverage the additional geometry revealed in non top-down views, we suggest a new, novel architecture: multi-view GQ-CNN (MV-GQ-CNN), which generates grasps that are agnostic to camera location. In this work we explore 6-DoF grasp planning by leveraging the fast 3D reconstruction from multiple RGB images a Learnt Stereo Machine \cite{kar2017learning} network achieves, along with the adaptation of a robust GQ-CNN \cite{mahler2016dex, mahler2017dex} grasping policy to non top-down views. This paper makes four contributions: (1) An adaptation of an LSM network that is trained on synthetic graspable objects, which are different from ShapeNet objects that the network was originally trained on. (2) An automated process to generate synthetic RGB and depth images from multiple view points to train the LSM network. (3) A multi-view grasp planning network, or MV-GQ-CNN, which is based on GQ-CNN and adapted to the images of objects from varying camera views. (4) An evaluation of the feasibility of depth maps generated from LSM for grasp planning.

\section{Related Works}

% \subsection{Structure From Motion (SfM)}
% In 3D reconstruction, Structure from Motion (SfM) is used when 3D point positions are not known in advance \cite{szeliski2010computer}. SfM simultaneously recovers the 3D structure and pose of the camera from image correspondences given multiple frames of RGB images. In this way, SfM estimates the 3D locations of points on the object’s notable geometric features from continuous frames. One limitation of SfM is that the objects being reconstructed must have notable geometric features such as contours, edges, and vertices. Thus, the objects need to be non-reflective and chromatic for the feature points to be detected and recorded. In addition, accurate SfM is computationally expensive and takes an infeasible amount of time for an online grasping task \cite{oliensis2000}.

\subsection{Multi-View Stereopsis (MVS)}

In 3D reconstruction, multi-view stereopsis (MVS) estimates dense representation from overlapping images \cite{zhang2016health, zhang2020dex, zhang2023flowbot++}. In other words, the goal of multi-view stereo is 3D reconstruction from multiple images taken from known camera viewpoints. While extremely accurate MVS algorithms exist, they are computationally expensive \cite{zhang2021robots, avigal20206, avigal2021avplug, devgon2020orienting}. For example, MVSNet, considered state-of-the-art in terms of accuracy and efficiency, takes 4.7 seconds per input view, which is an infeasible runtime for an online grasping task utilizing multiple views \cite{MVS, pan2023tax, eisner2022flowbot3d}.

\subsection{Learnt Stereo Machine (LSM)}

Learnt Stereo Machines (LSM) solve the problem of multi-view stereopsis through using a learnt system that leverages feature projection and unprojection along viewing rays for 3D reconstruction \cite{elmquist2022art, sim2019personalization}. LSM allows for reconstruction from much fewer images, as well as a near-instant 3D reconstruction after the network has been trained, due to its feed-forward nature \cite{kar2017learning, lim2021planar, lim2022real2sim2real}.

\subsection{Grasp Planning}
Robotic grasping is one of the most widely explored areas of motion planning and manipulation. Data driven approaches, such as Dex-Net, learn from a large dataset of grasps and manage to generalize to grasp unseen objects. Specifically, Dex-Net utilizes a built-in deep neural network architecture called Grasp Quality CNN (GQ-CNN), which was trained on millions of top-down depth images and takes in a top-down depth map of the object of interest and sample all possible grasps using Cross Entropy Methods (CEM) and picks the best one according to its confidence level and grasp robustness \cite{mahler2016dex, mahler2017dex}.

To address the large state space and the huge number of possible grasps, most data driven approaches limit their grasps to 4-DoF crane grasps, where the gripper approaches the object in a top-down manner. Although it makes the problem of robust grasping more feasible, it restricts the variety of grasps that can be sampled, and the range of objects or scenarios that can be handled by the algorithm, and therefore it is not good enough for the home environment. Morrison et al. explored a closed loop approach, where the grasp is planned in a 4-DoF setting, but better grasps can be found as the arm approaches to grasp the object ~\cite{morrison2018closing, jin2024multi, shen2024diffclip, yao2023apla}. In contrast to the top-down 4-DoF grasp setting, 6-DoF grasp planning is a grasping technique that is not limited by the depth image plane. Most of the existing grasp planning systems represent the grasp as an oriented rectangle in the image. This 3-DoF representation constrains the gripper pose to be parallel to the image plane. The drawbacks of 3-DoF grasping are apparent: it limits grasp diversity, and the robotic arm might interfere with the planned grasp if the grasp is generated using a single depth image. Static top-down depth cameras also severely constrain the setup of the robot's work-space. Mousavian et al. proposed a VAE-based 6-DoF grasp generator that can efficiently sample grasps and train the robot \cite{mousavian2019}. Zhang et al. introduced a point cloud-based 6-DoF grasp planning tool using mobile phone augmented-reality \cite{zhangdex}. Staub et al. proposed a 6-DoF grasp planning technique on Toyota HSR robots \cite{staub2019dex}.

\section{Problem Statement}

The goal of the paper is to map multiple RGB images of the same scene from differing points of view to an optimal 6-DoF grasp. Let $\mathbf{g} = (\mathbf{p}, \phi, \theta)$ denote a parallel-jaw grasp, where $\mathbf{p} = (x,y,z) \in \mathbb{R}^3$ is the gripper location in 3D space, $\phi \in \mathbb{S}^1$ is the angle in the table plane and $\theta \in \mathbb{S}^1$ is the angle between the camera optical axis the table plane. Given $n$ RGB images $X_1\dots X_n$ and their associated camera parameters $C_1\dots C_n$, reconstruct 3D geometry using an LSM which yields depth maps $D_1\dots D_n$,
plan grasps on $D_1\dots D_n$ using a multi-view GQ-CNN (MV-GQ-CNN) grasping policy, and return $\textbf{g}^*$, the grasp with the highest grasp quality value.  

\section{Methodology}

\subsection{Generating Depth Maps with Pretrained LSM}

\begin{figure}[]
    \centering
    \includegraphics[scale=0.5]{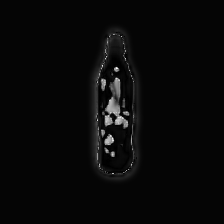}
    \caption{\textbf{Pretrained Reconstruction Error}: Pixel-wise difference between the LSM-predicted depth map and the ground truth depth map shows poor-quality prediction of a graspable bottle's depth image.}
    \label{fig:bad}
\end{figure}

LSM was trained and evaluated on Shapenet, whose categories mostly consist of cars, aeroplanes, tables, chairs, etc. This distribution of objects is vastly different from one consisting of graspable objects. Fig. \ref{fig:bad} shows the poor performance of this pretrained network on a simple, graspable object. This motivates retraining the LSM on new data. 

\subsection{Dataset Generation}
Since the original LSM was trained on ShapeNet, which is comprised of non-graspable objects \cite{kar2017learning}, we decide to retrain the network using data from graspable objects. 

To achieve this, we generate synthetic training data for LSM. We first sample category-specific meshes from a graspable objects' meshes dataset, which will result in a dataset of meshes of a number of objects from the same category. Afterwards, we load the meshes to PyRender, and apply randomized vertex and face colors to each mesh. To render images, we sample 20 different cameras from a hemisphere, where the camera angles are specified by the extrinsics $R$ and $t$. For the camera intrinsics parameter, we use the intrinsics of a PhoXi depth camera, such that each output depth image is effectively taken by a synthetic PhoXi depth camera, which was the camera we used when we trained Dex-Net. Using the 20 different camera views, we render 20 synthetic RGB and depth images of the object in the scene.

We also have a separate process for generating the testing data. Specifically, we sample meshes from the same category and apply novel features and coloring to the meshes, and then render 10 RGB and depth images from 10 novel camera views in order to test the performance of the network. Among the 10 novel camera views, one top-down angle is always retained because we are most interested in that orientation.

\begin{figure}
    \centering
    \includegraphics[scale=0.3]{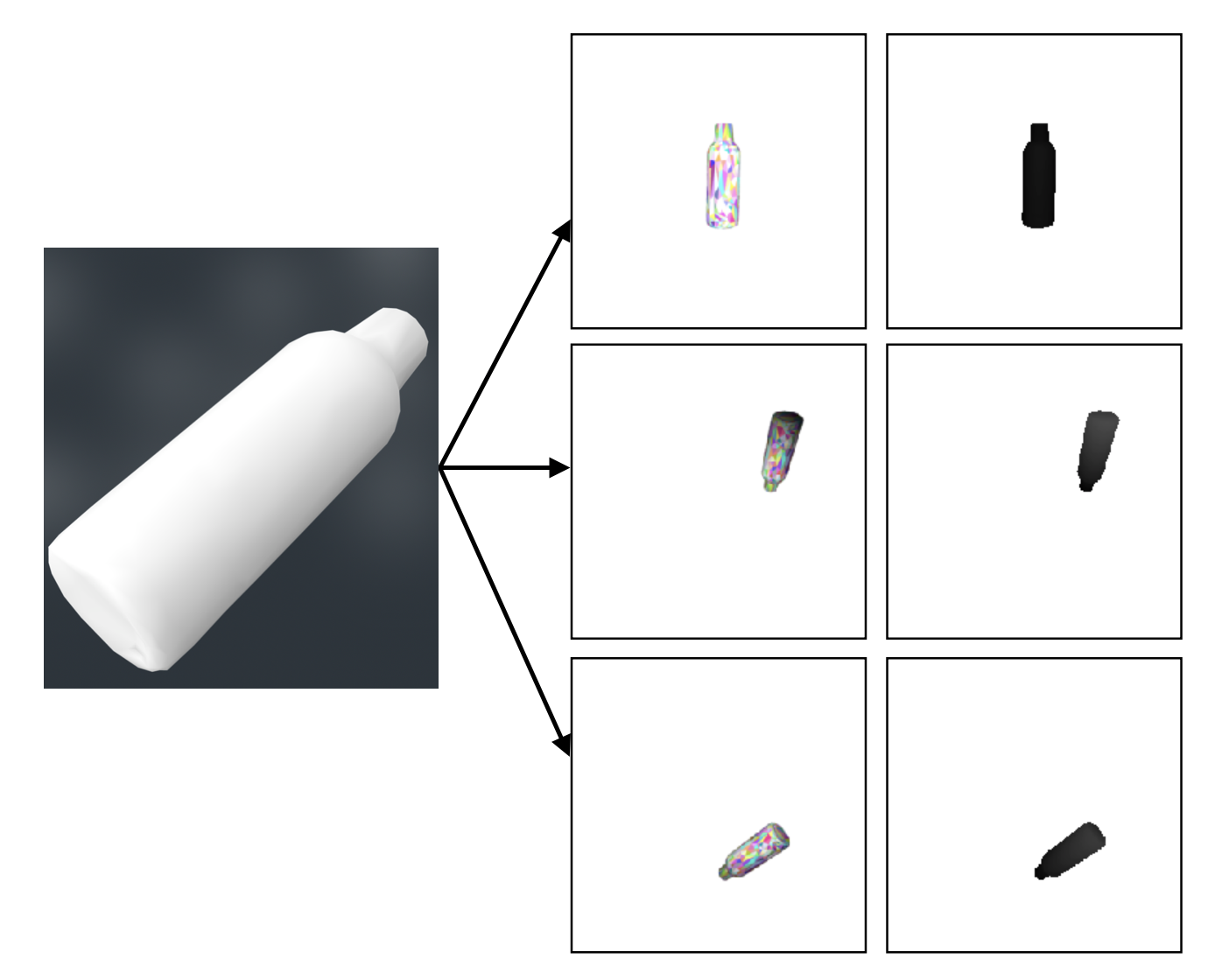}
    \caption{\textbf{Data Generation Illustrations}: \textit{Left:} the original bottle object 3D mesh. \textit{Middle:} renderings of 3 RGB images from 3 different angles. \textit{Right:} renderings of the 3 corresponding depth maps.}
    \label{fig:datagen}
\end{figure}

Some examples from our data generation process are shown in Fig. \ref{fig:datagen}.

\subsection{Generating Depth Maps with Retrained LSM}

We train on images from 20 cameras fixed in space. Thus, for each object, we have 20 RGB images, 20 depth images, and 20 cameras. We train an LSM that minimizes the distance between the ground truth depth map and the predicted depth map. Each input includes the top-down view, as well as 3 additional randomly sampled views. We train using 20000 iterations, using 31 bottles as training data, and 4 bottles as validation data. Once trained, to generate depth maps $D_1\dots D_n$ using an LSM, we feed in $X_1\dots X_n$ and $C_1\dots C_n$ into the LSM. 

\subsection{Grasp Analysis of the Generated Depth Maps}
We generate a new synthetic set of RGB images and their corresponding depth images following the data generation procedure described in 4.2, so that the new set is different from the images used for training and validation. We feed the color images into the retrained LSM and obtain a set of predicted depth images. The predicted depth images have 3 identical channels with identical integer pixel values varying in the range of 0-255 across the channels, so we slice a single channel and normalize it to a meter scale, specified by $z_\text{near}$ and $z_\text{far}$ of the depth camera. In addition, we train a GQ-CNN grasp planning policy following the training procedure described in~\cite{mahler2017dex}. We then pass both the rendered ground truth depth images and the predicted depth images to a GQ-CNN policy, and compare the grasp quality of each to determine whether or not the predicted images can be used for grasping. 
\begin{figure}
    \centering
    \begin{tabular}{c@{\hspace{5pt}}c}
        \frame{\includegraphics[width=.4\columnwidth, height=34.3mm]{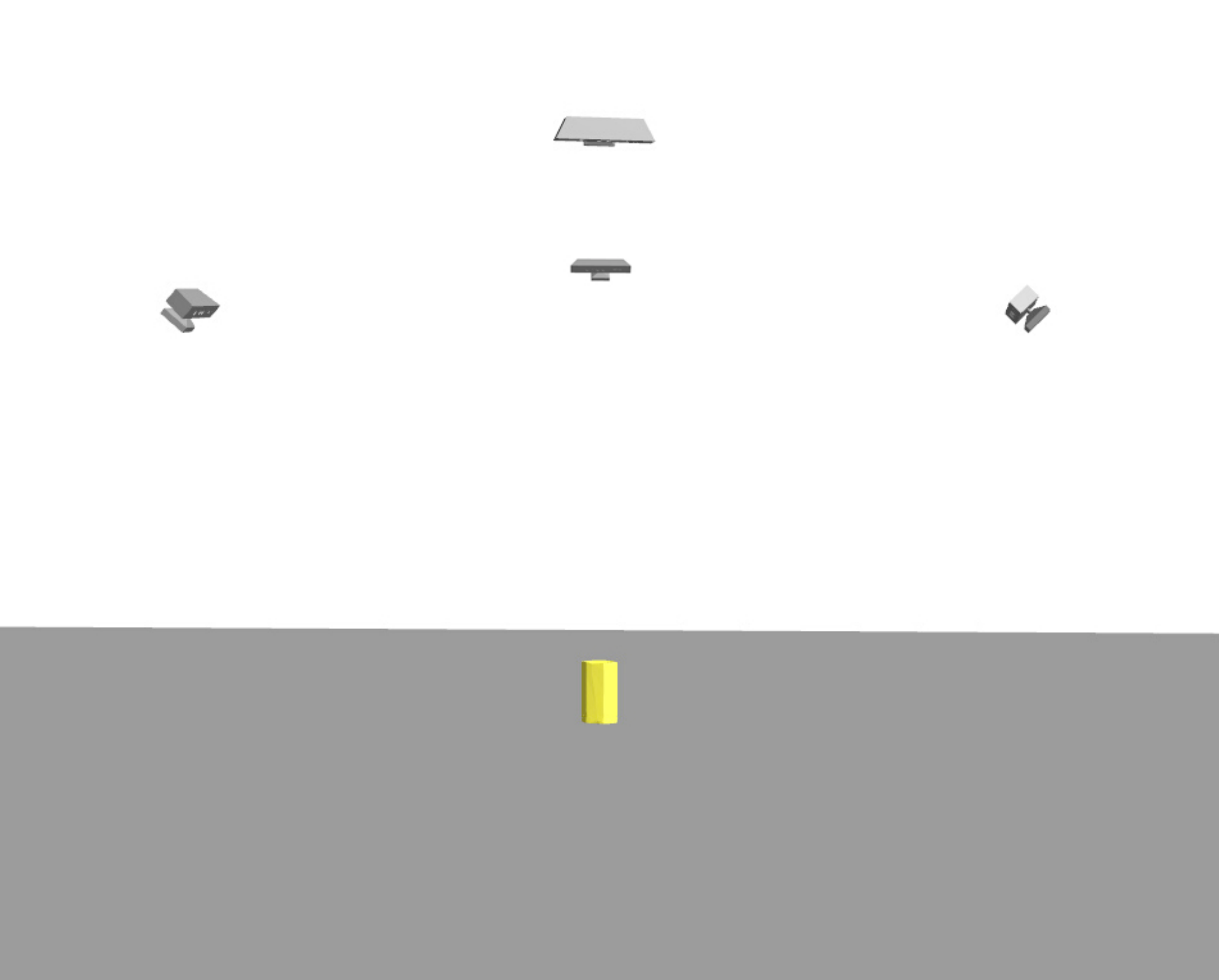}} & \frame{\includegraphics[width=.4\columnwidth]{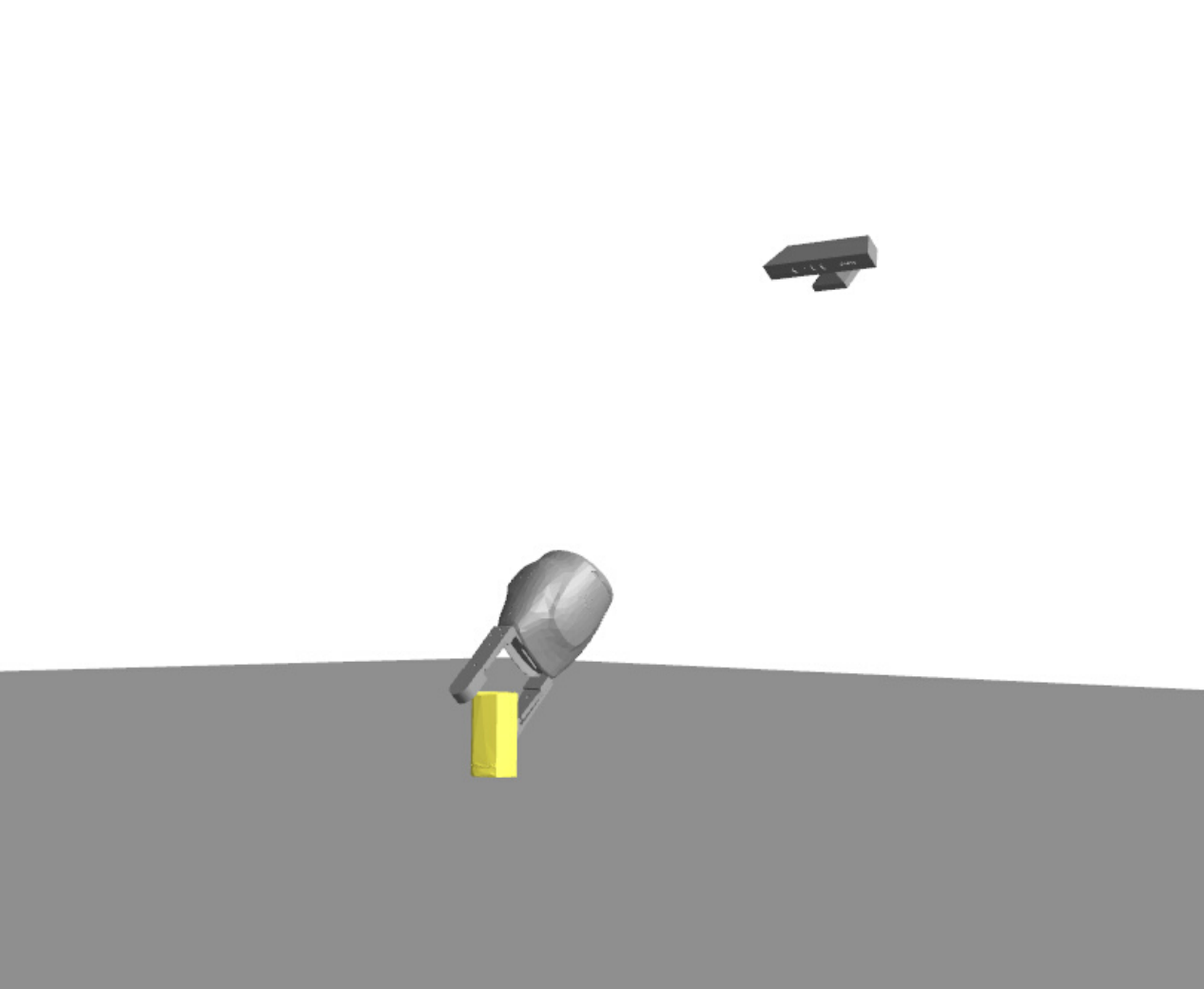}} \\
    \end{tabular}
    \caption{\textbf{Multi-View Grasping}: Cameras are located in different locations in a hemisphere around the target object. A grasp is planned from each view, and the grasp with the highest success probability is chosen and applied.}
\end{figure}

\subsection{Multi-view Grasping}
To fully leverage the information distilled in the 3D representation generated using the LSM to plan 6-DoF grasps in the 3D space, we extend the work in~\cite{mahler2017dex} that uses a GQ-CNN grasping policy in a top-down setting to a multi-view setting. In this new setting, the object lies in the center of a hemisphere and the camera can be located at any point in the hemisphere, oriented such that its optical axis is directed at the object. We train the MV-GQ-CNN policy on a dataset of synthetic point clouds, following the training procedure described in~\cite{mahler2017dex}. We vary the camera location and orientation, as opposed to the original setting in which the camera was fixed approximately 1 meter above the object, while the camera's optical axis merges with the table normal. We take images of the target object by fixing the camera in 5 locations on the circumference of the hemisphere, as seen in Fig. 4. We then compare the grasp quality of the best grasp found from each camera view, and return the one with the highest grasp quality.

\section{Results}

\subsection{LSM}

\begin{figure}[ht]
\begin{center}
\begin{tabular}{c|c}
\hline
Pretrained Network & Our Network \\
\hline\hline
\vspace{1pt} & \vspace{1pt} \\ 
\includegraphics[width=.352\linewidth]{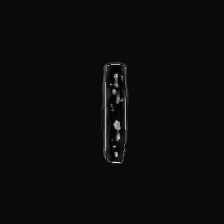} & \includegraphics[width=.352\linewidth]{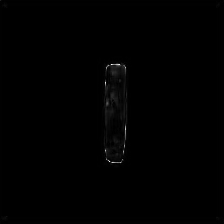} \\
\includegraphics[width=.352\linewidth]{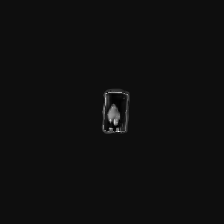} & \includegraphics[width=.352\linewidth]{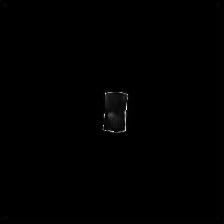} \\
\hline
\vspace{1pt} & \vspace{1pt} \\ 
\includegraphics[width=.352\linewidth]{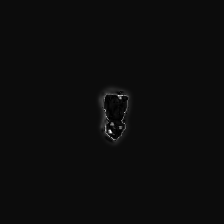} & \includegraphics[width=.352\linewidth]{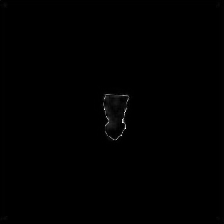} \\
\includegraphics[width=.352\linewidth]{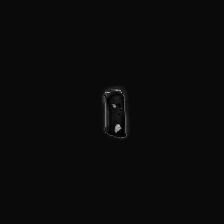} & \includegraphics[width=.352\linewidth]{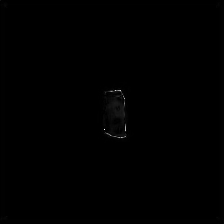} \\
\end{tabular}
\end{center}
\caption{\textbf{Unseen Objects}: Difference between ground truth and predicted depth map. Top half contain unseen bottles, bottom half contains objects from unseen categories. Our network has better performance on both types of unseen objects, as these graspable objects better match our training distribution. From top to bottom: shampoo bottle, waffleroll, sitting cat, NutellaGo package.}
\label{fig:unseen1}
\end{figure}

Our LSM trained on our generated data performs well on seen and unseen data, outperforming the LSM trained on ShapeNet on unseen bottles, as well as other graspable objects. In terms of unseen bottles, the outputted depth maps resemble the ground truth, as seen in Fig. \ref{fig:unseen1}. We see our LSM is able to generalize to unseen graspable objects relatively well, with minimal noise in the depth map, also seen in Fig. \ref{fig:unseen1}. For top-down views of bottles, one shot reconstruction is sufficient, as the mean pixel-wise reconstruction error remains similar across the dataset, regardless of the number of inputs used (Table 1).

\begin{table}[ht!]
\begin{center}
\begin{tabular}{|l|c|c|c|}
\multicolumn{1}{c}{} & 
\multicolumn{1}{c}{\includegraphics[width=.22\linewidth]{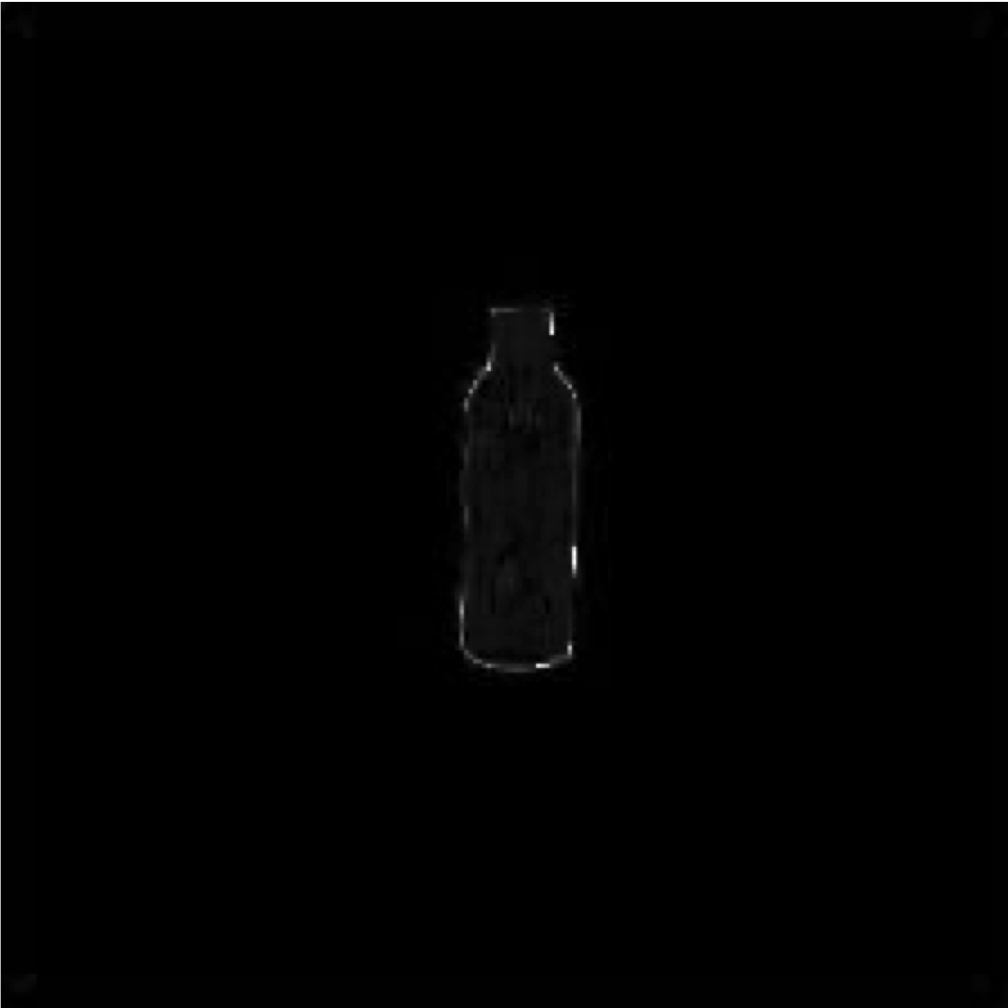}} &
\multicolumn{1}{c}{\includegraphics[width=.22\linewidth]{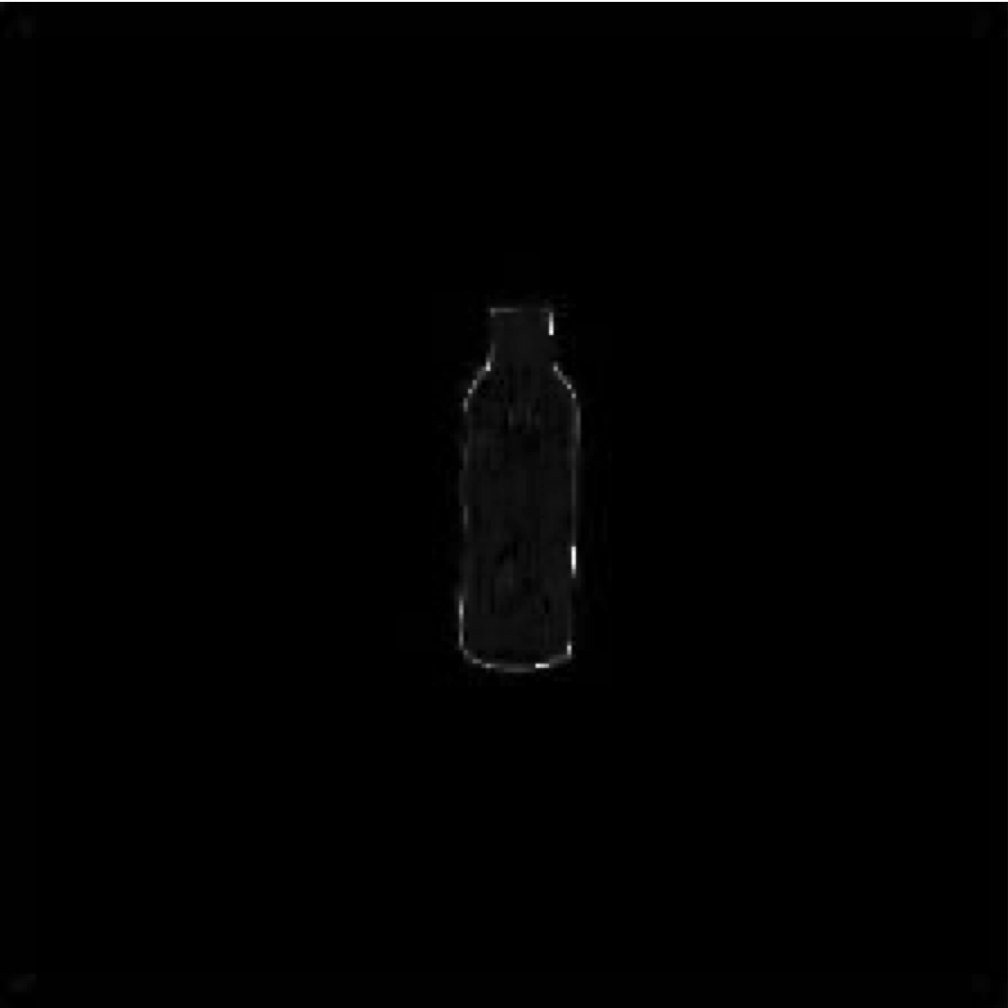}} &
\multicolumn{1}{c}{\includegraphics[width=.22\linewidth]{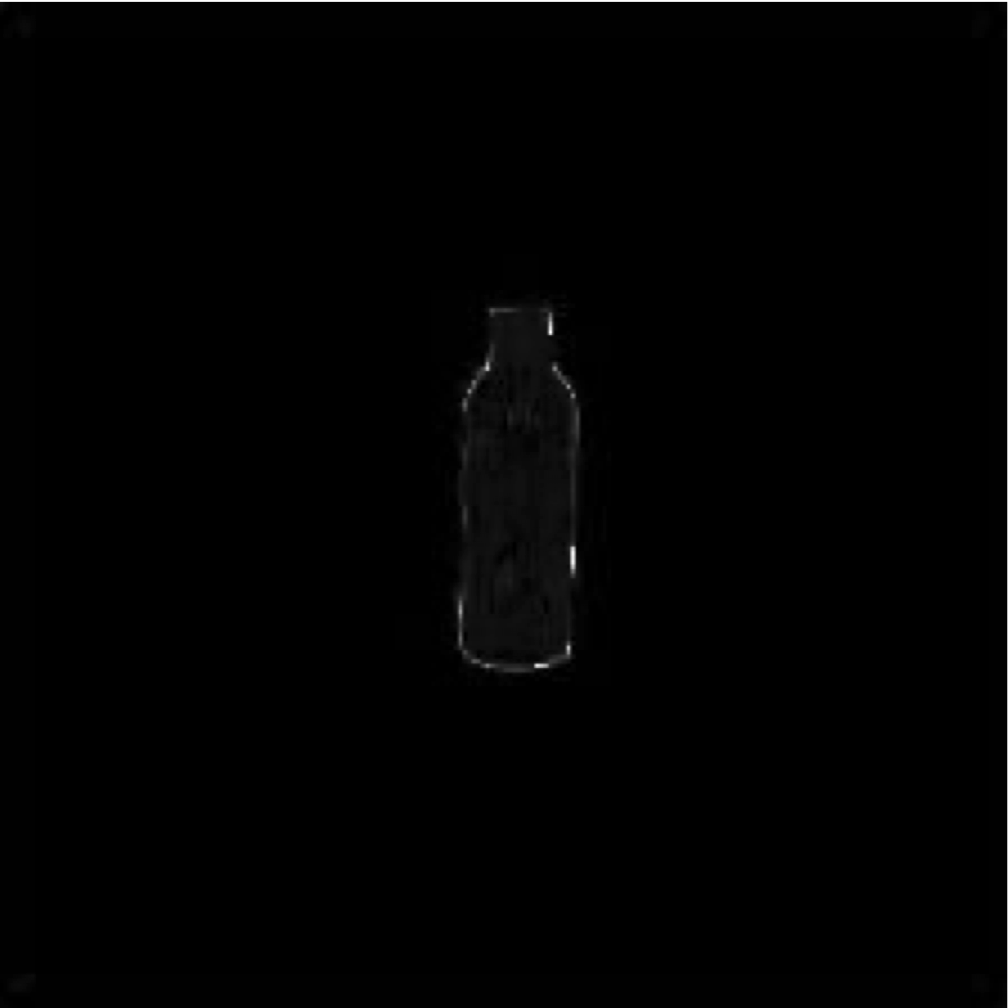}} \\
\hline
& 1-Shot & 3-Shot & 9-Shot \\
\hline\hline
Above & \textbf{0.003496} & 0.003527 & 0.003527 \\
\hline
Dataset & \textbf{0.002885} & 0.002896 & 0.002896 \\
\hline 
\end{tabular}
\end{center}
\caption{\textbf{Reconstruction Error:} Absolute sum of differences between predicted depth map and ground truth. $x$-shot refers to inputting $x$ unique views into the LSM. We see that for top-down views of bottles, one-shot reconstruction is sufficient, as the pixel-wise reconstruction error remains similar across the dataset.}
\end{table}

\begin{figure*}
    \centering
    \vspace{0.3cm}
    \begin{tabular}{c@{\hspace{2.5pt}}c@{\hspace{2.5pt}}c@{\hspace{2.5pt}}c@{\hspace{2.5pt}}c@{\hspace{2.5pt}}c@{\hspace{2.5pt}}}
    
    Bowl&Toy-Chair&Mug&Squash&Hole-Puncher&Toy-Car\\
    \includegraphics[width=22mm, height=18.3mm]{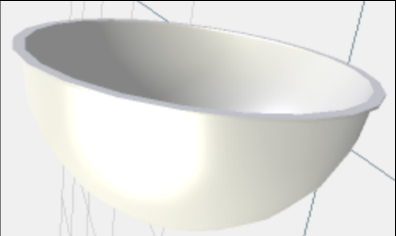}&
    \includegraphics[width=22mm, height=18.3mm]{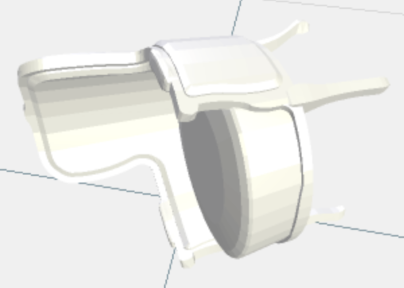}&
    \includegraphics[width=22mm, height=18.3mm]{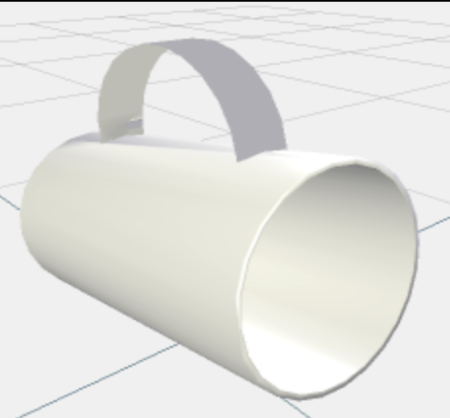} &
    \includegraphics[width=22mm, height=18.3mm]{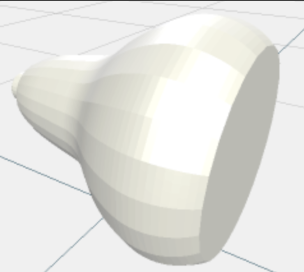} &
    \includegraphics[width=22mm, height=18.3mm]{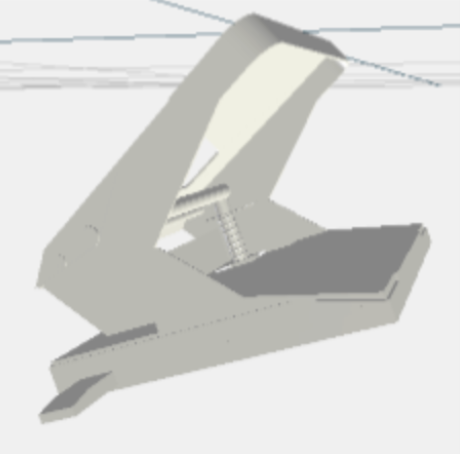}&
    \includegraphics[width=22mm, height=18.3mm]{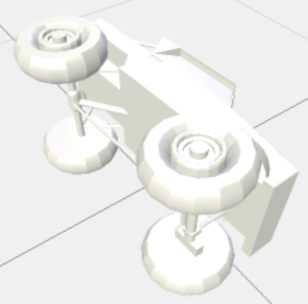}\\
    \includegraphics[width=22mm, height=18.3mm]{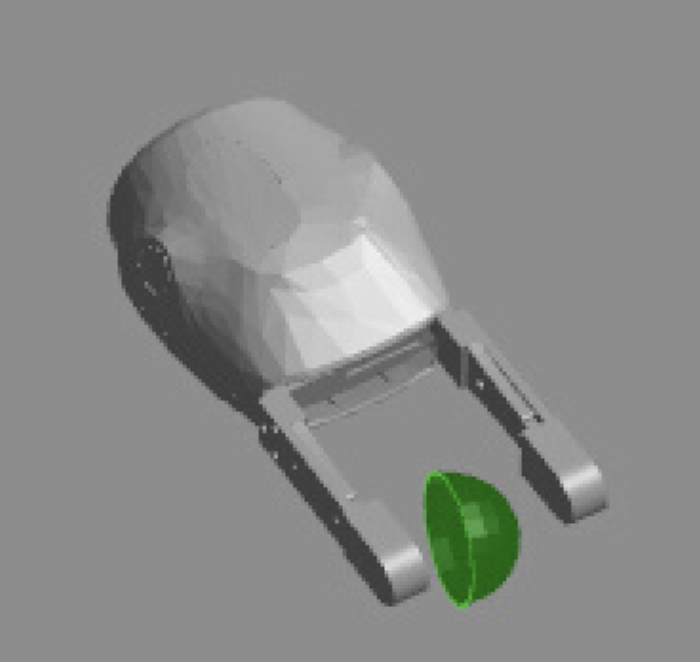}&
    \includegraphics[width=22mm, height=18.3mm]{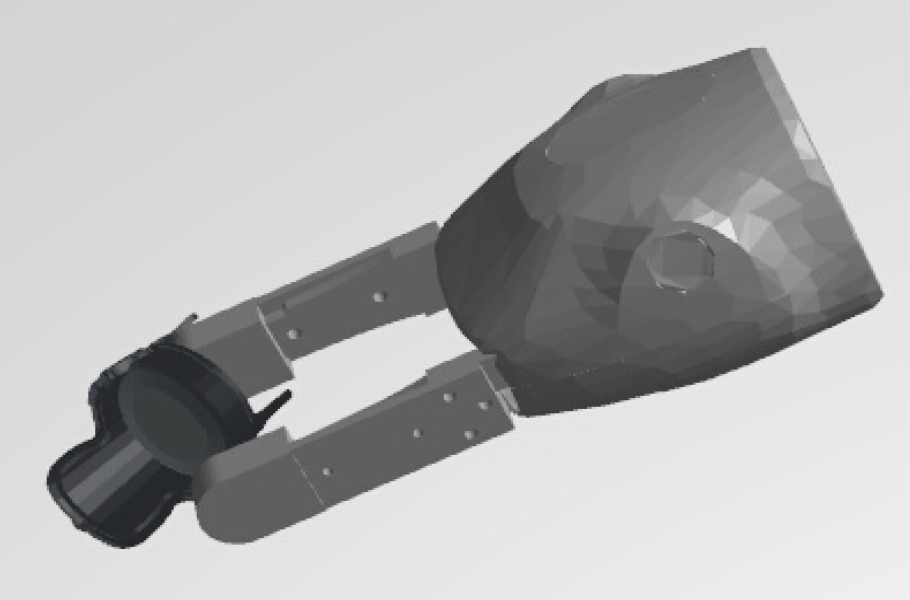}&
    \includegraphics[width=22mm, height=18.3mm]{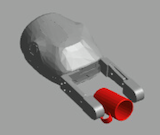} &
    \includegraphics[width=22mm, height=18.3mm]{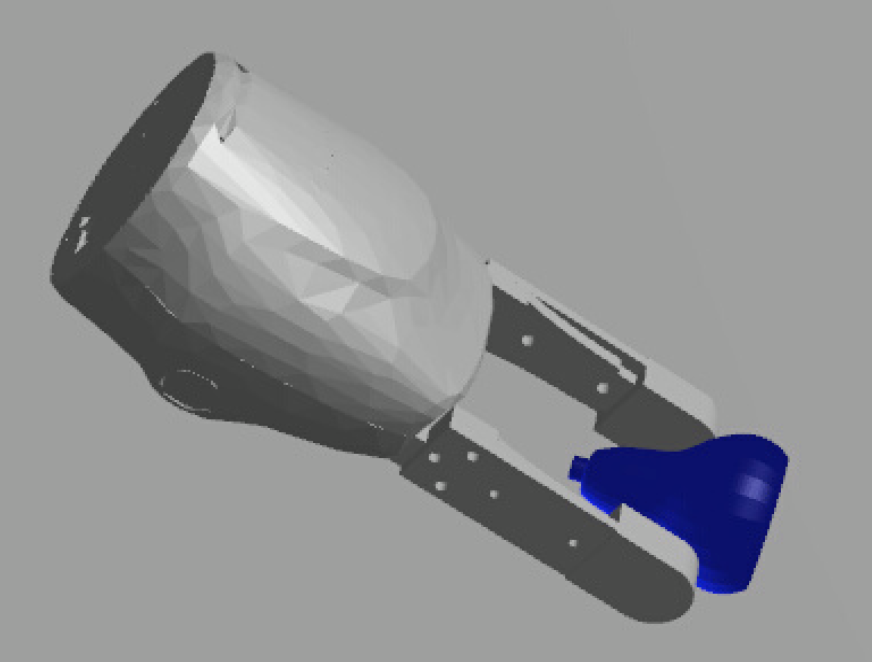} &
    \includegraphics[width=22mm, height=18.3mm]{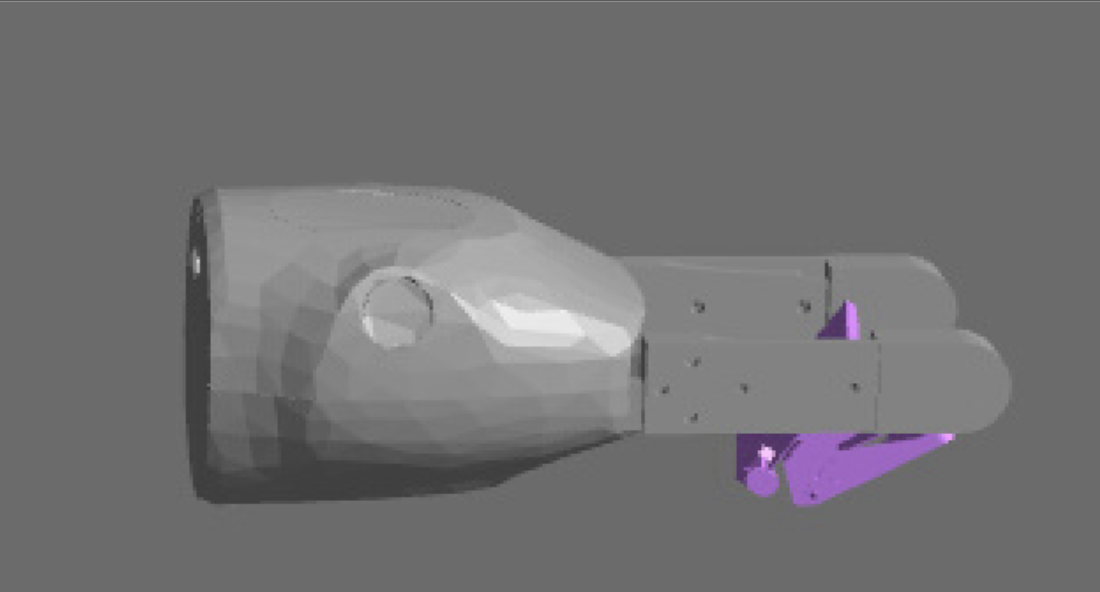}&
    \includegraphics[width=22mm, height=18.3mm]{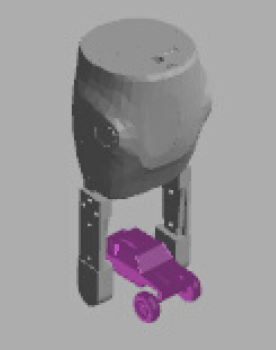}\\
    $Q = 0.96$&$Q = 0.6$&$ Q = 0.89$&$ Q = 0.96$&$ Q = 0.77$&$ Q = 0.89 $
     
    \end{tabular}

    \caption{\textbf{MV-GQ-CNN Renderings}: Three dimensional renderings of the grasps using MV-GQ-CNN on 6 different objects. The values in the bottom row are the Q-values of the grasps, representing how confident the network is when planning the grasps.}
    \label{fig:mvgqcnn}
\end{figure*}

% \begin{figure}[ht]
% \begin{center}
% \begin{tabular}{c|c}
% \hline
% Pretrained Network & Our Network \\
% \hline\hline
%  & \\ 
% \includegraphics[width=.4\linewidth]{figs/diff_CatSitting_800_tex_pre.png} & \includegraphics[width=.4\linewidth]{figs/diff_CatSitting_800_tex.png} \\
% \includegraphics[width=.4\linewidth]{figs/diff_NutellaGo_800_tex_pre.png} & \includegraphics[width=.4\linewidth]{figs/diff_NutellaGo_800_tex.png} \\
% \end{tabular}
% \end{center}
% \caption{\textbf{Unseen Categories}: Difference between ground truth and predicted depth map for unseen categories for both network: sitting cat (top), and NutellaGo package (bottom).  Our network has better performance on these unseen categories, as it better matches the distribution.}
% \label{fig:unseen2}
% \end{figure}

\begin{figure}
\begin{center}
\begin{tabular}{c|c}
\hline
\footnotesize{Grasps via Ground Truth Images} & \small{Grasps via Predicted Images} \\
\hline\hline
\vspace{-4pt} & \vspace{-4pt} \\ 
\includegraphics[width=.4\linewidth]{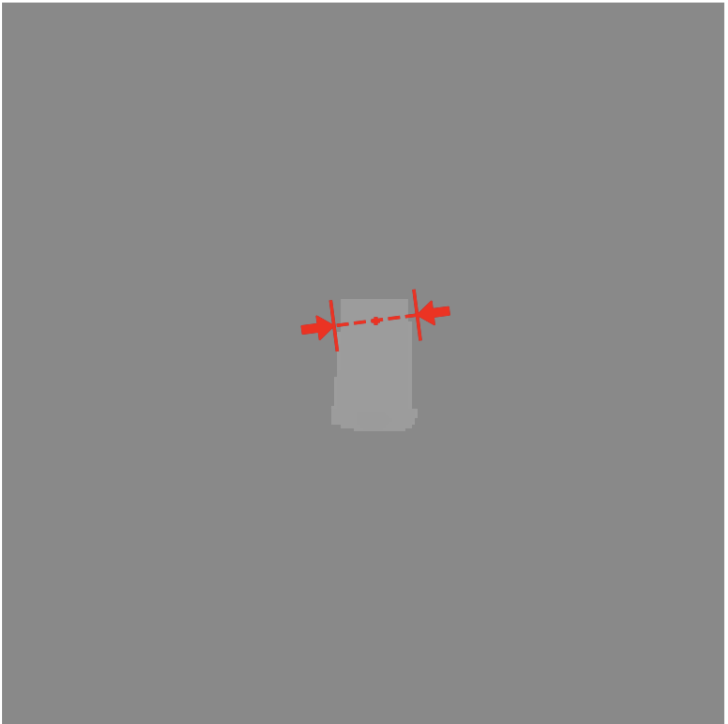} & \includegraphics[width=.4\linewidth]{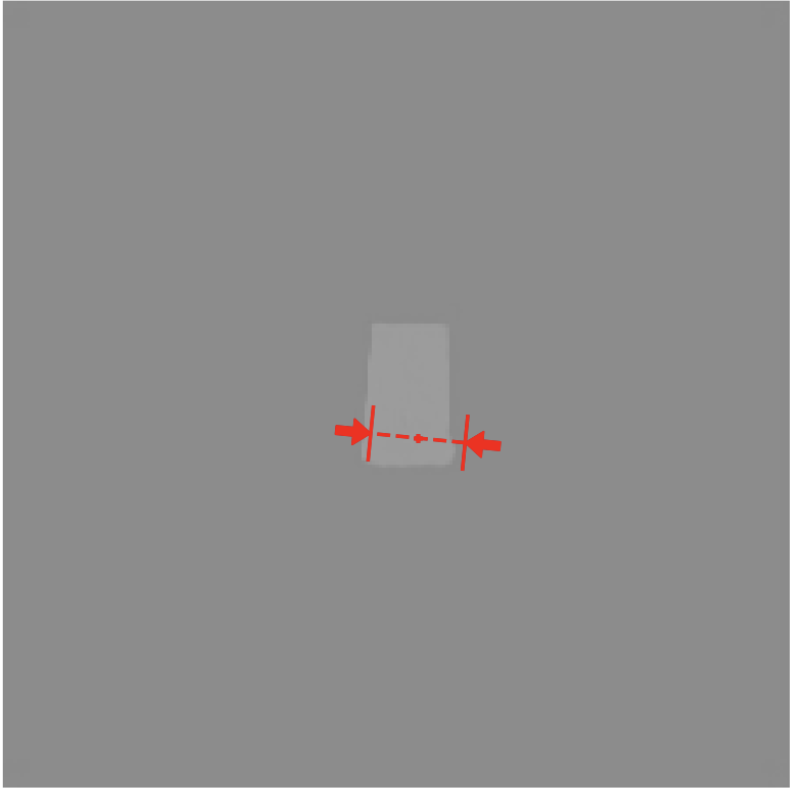} \\
\includegraphics[width=.4\linewidth]{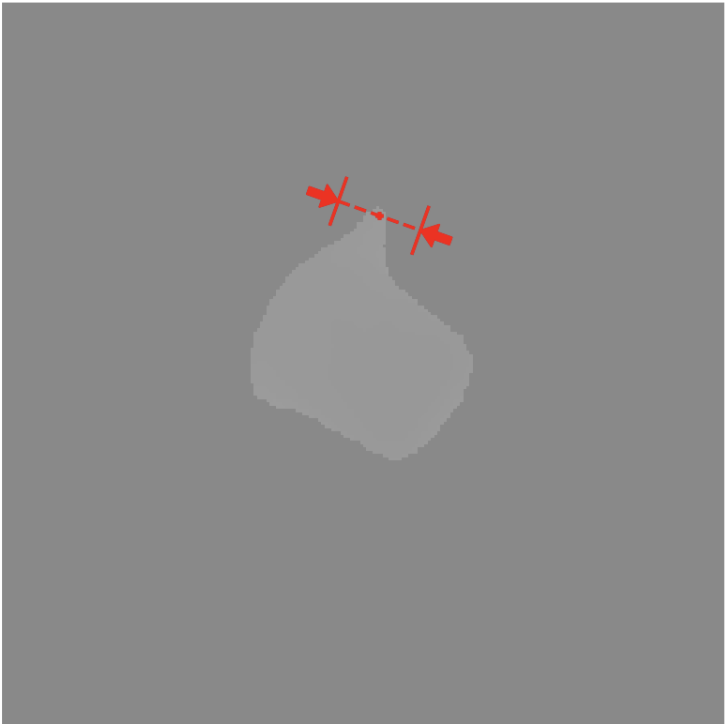} & \includegraphics[width=.4\linewidth]{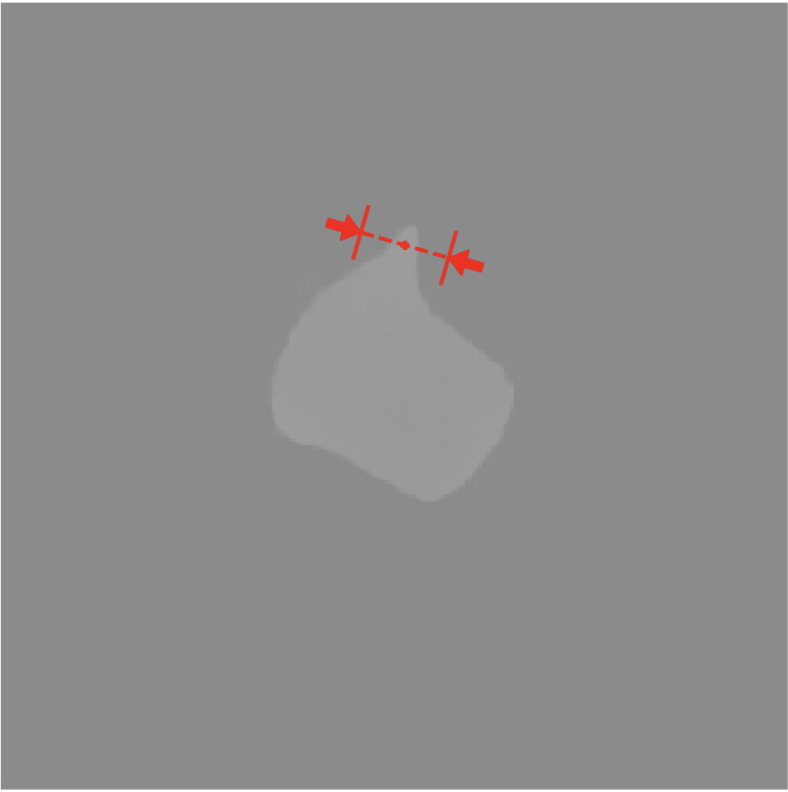} \\
\end{tabular}
\end{center}
\caption{\textbf{GQ-CNN Grasping Results}: Difference between grasps from ground truth images versus LSM-predicted images. While visually similar, the top-left grasp has a Q-value of 0.652, the top-right grasp has a Q-value of 0.830, the bottom-left grasp has a Q-value of 0.300, and the bottom-right grasp has a Q-value of 0.080.}
\label{fig:gqcnn}
\end{figure}
\subsection{Grasp Analysis Using GQ-CNN}
Using the original GQ-CNN trained on top-down views, we compare the grasp quality results of GQ-CNN run on the rendered ground truth depth images and on the LSM-predicted depth images. Both sets of the images are normalized to the same depth range in meters, specified by the synthetic PhoXi camera's $z_\text{near}$ and $z_\text{far}$. In Fig. \ref{fig:gqcnn}, we show comparisons between the grasp quality of the best grasp found on top-down ground truth depth maps and on top-down view LSM-predicted depth maps. As seen in the examples in Fig. \ref{fig:gqcnn}, although the grasps often look similar between predicted and ground truth images, their are some differences in terms of the confidence level of the GQ-CNN network about the grasps.

We compare the grasp quality of the best grasp found from different camera views on objects with different topologies and in different poses, in order to exploit the multiple views to find the best grasp. We use a set of 3D objects that can be found in the home environment: a bowl, toy-chair, mug, squash, hole-puncher and toy-car. For each object we present the best grasp quality out of 5 different views, determined by the MV-GQ-CNN grasp planning policy, compared with the grasp quality from a top-down view. The results are presented in Table 2. The experiments suggest that planning a grasp using different views of the target object can outperform the traditional top-down method. To fully evaluate these results we would need to get access to the physical robotic platform and conduct the experiments with the real objects, however this is will be left for future work, once the shelter-in-place ends.

\begin{table}[hbt]
\begin{center}
\begin{tabular}{|l|c|c|}
    \hline
    \textbf{Object} & \textbf{Top-Down} & \textbf{Multi-View} \\ \hline
    Chair           & $N/A^\dagger$    & \textbf{0.60}        \\ \hline
    Mug             & \textbf{0.89}    & 0.84                 \\ \hline
    Truck           & \textbf{0.89}    & 0.72                 \\ \hline
    Hole Punch      & 0.62             & \textbf{0.77}        \\ \hline
    Bowl            & 0.87             & \textbf{0.96}        \\ \hline
    Squash          & 0.90             & \textbf{0.96}        \\ \hline
    \multicolumn{3}{l}{\small $\dagger$ Grasp not found, often due to failing to segment edge pixels}\\
\end{tabular}
\end{center}
\caption{\textbf{Grasping From Different Views}: A comparison of grasp qualities planned from a top-down view and from multiple views, on objects that can be found in the home environment and have complex topology.}
\end{table}

\section{Conclusion}
We present a novel 6-DoF grasp planning tool that combines Learnt Stereo Machine (LSM), a deep learning-based multi-view stereopsis technique, with Dex-Net, a state-of-the-art 4-DoF grasp planning system. We retrain the LSM using an automated synthetic RGB and depth images generation process. We obtain decent grasp planning results which suggests that the depth images predicted using LSM can be used for grasping. We also train a Multi-View GQ-CNN and show it can outperform the original top-down grasp planner. In the future, we plan to fully integrate the MV-GQ-CNN with the LSM, and to evaluate the performance of the system on a physical robot. Furthermore, we would also try to modify LSM accordingly to reconstruct a cluster of objects in a scene to address objects in clutter.

% refs on new column
% \vfill\null
% \columnbreak

{\small
\bibliographystyle{ieee_fullname}
\bibliography{egbib}
}

\end{document}